\pdfoutput=1

\documentclass[11pt]{article}

\usepackage[]{acl}

\usepackage{times}
\usepackage{latexsym}

\usepackage[T1]{fontenc}

\usepackage[utf8]{inputenc}

\usepackage{microtype}

\usepackage{inconsolata}
\usepackage{tabularx}
\usepackage{booktabs}
\usepackage[ruled,linesnumbered]{algorithm2e}
\usepackage{amsfonts,amssymb}
\usepackage{graphicx}
\usepackage{multirow}
\usepackage{longtable}
\usepackage{booktabs}
\usepackage{color}
\usepackage{makecell}

\title{InsCL: A Data-efficient Continual Learning Paradigm for Fine-tuning \\
Large Language Models with Instructions}

\author{
Yifan Wang$^{1*}$, Yafei Liu$^{2}\Thanks{\hspace{1mm}Equal contribution.}$  , Chufan Shi$^{1}$, Haoling Li$^{1}$,\\
{\bf Chen Chen$^{2}$, Haonan Lu$^{2}$, Yujiu Yang$^{1}$\Thanks{\hspace{1mm}Corresponding author.}} \\
$^1$ Tsinghua University \quad $^2$ OPPO AI Center\\
\texttt{\{wangyifa22,scf22,li-hl23\}@mails.tsinghua.edu.cn}\\
\texttt{\{liuyafei,chenchen4,luhaonan\}@oppo.com} \quad \texttt{yang.yujiu@sz.tsinghua.edu.cn}
}

\begin{document}
\maketitle
\begin{abstract}
Instruction tuning effectively optimizes Large Language Models (LLMs) for downstream tasks. 
Due to the changing environment in real-life applications, LLMs necessitate continual task-specific adaptation without catastrophic forgetting. 
Considering the heavy computational cost, replay-based Continual Learning (CL) methods are the simplest and most widely used for LLMs to address the forgetting issue.
However, traditional replay-based methods do not fully utilize instructions to customize the replay strategy.
In this work, we propose a novel paradigm called Instruction-based Continual Learning (InsCL).
InsCL dynamically replays previous data based on task similarity, calculated by Wasserstein Distance with instructions.
Moreover, we further introduce an Instruction Information Metric (InsInfo) to quantify the complexity and diversity of instructions. 
According to InsInfo, InsCL guides the replay process more inclined to high-quality data.
We conduct extensive experiments over 16 tasks with different training orders, observing consistent performance improvements of InsCL.
When all tasks have been trained, InsCL achieves performance gains of 3.0 Relative Gain compared with Random Replay, and 27.96 Relative Gain compared with No Replay.

\end{abstract}

\section{Introduction}

\begin{figure}[t]
   \centering
   \includegraphics[width=0.835\linewidth]{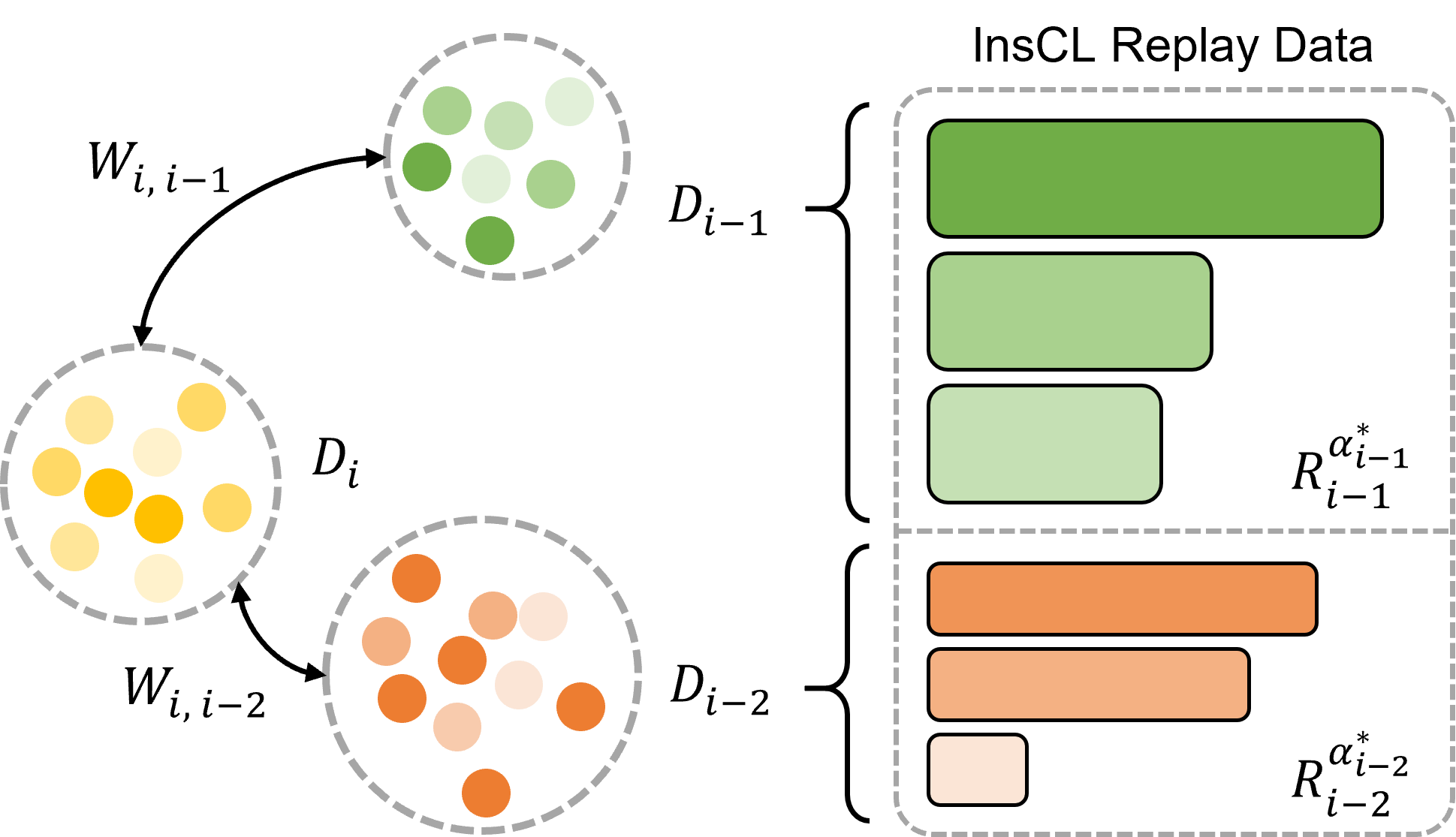}
   \caption{
   The framework of InsCL, the index denotes task id.
   $D$ represents task data, and $R$ represents the sampled data to replay.
   InsCL dynamically replays $\alpha^*$ data for each previous task based on the task similarity calculated via Wasserstein Distance $W$.
   The dots represent instructions included in each task, and the darker colors represent higher InsInfo.
   The size of each color bar denotes the corresponding amount of replay data.
   }
   \label{fig:InsCL}
\end{figure}

Large Language Models (LLMs) show remarkable capabilities from a wide range of Natural Language Processing (NLP) tasks~\cite{brown2020language, ouyang2022training, touvron2023llama}, demonstrating large potential in handling various task-specific settings. 
%
To complete realistic downstream tasks, recent works suggest that instruction tuning is an incredible method for unleashing the power of LLMs~\cite{wei2021finetuned, peng2023instruction, shi2023specialist}.
However, in real-life applications, the consistent emergence of new corpora and knowledge changes task schemas frequently, necessitating continual task-specific adaptation for LLMs~\cite{jin2021lifelong, daruna2021continual}.
%
Accordingly, Continual Learning (CL) is proposed to learn a sequence of tasks incrementally, updating models for the changing environment without catastrophic forgetting~\cite{goodfellow2013empirical, kemker2018measuring}. 

Considering the heavy burden on computing time and GPU memory of tuning LLMs, replay-based methods are the simplest and most effective among all traditional CL methods.
Despite several replay-based methods that have been well-studied~\cite{ sun2019lamol, wang2020efficient, mi2020continual, qin2022elle}, some traditional strategies cannot achieve optimal performance in continual instruction tuning due to the unique data composition.
To address this issue, we propose a data-efficient paradigm called \textbf{Ins}truction-based \textbf{C}ontinual \textbf{L}earning (InsCL), applied to continual fine-tuning LLMs with natural language instructions.
InsCL effectively utilizes instructions as high-quality task descriptions, designing a dynamic instruction-information-based replay method.
As shown in Figure~\ref{fig:InsCL}, when the new task $D_i$ comes, InsCL will sample replay data $R$ from all the previous tasks
(here we list two previous tasks in Figure~\ref{fig:InsCL}).

InsCL dynamically replays $\alpha^*$ data from previous tasks based on their similarity with the current task.
We draw on the application of Optimal Transport~\cite{torres2021survey} in comparing different distributions and adopt Wasserstein Distance~\cite{liu2022wasserstein} as a similarity measure.
Since instructions naturally contain high-quality task-related descriptions, we use instructions to calculate Wasserstein Distance instead of using the full amount of data, significantly reducing the computational cost~\cite{cuturi2013sinkhorn}.
For the previous tasks that are more different from the current task, InsCL allocates a larger replay scale (\textbf{larger bar width} in Figure~\ref{fig:InsCL}).

After determining the sample size based on task similarity, InsCL leverages instruction information to guide the sampling process more inclined to high-quality data.
Prior works have shown that the performance with less but high-quality data can be comparable with full data~\cite{toneva2018empirical, abbas2023semdedup, tirumala2023d4}.
For instruction tuning scenarios, early attempts~\cite{wang2022self, xu2023wizardlm, ding2023enhancing} affirm that LLMs' performance can be improved by increasing the training template complexity and diversity.
Inspired by this, we propose an Instruction Information Metric (InsInfo) to quantify the complexity and diversity of instructions.
With InsInfo-guided sampling, InsCL replays more high-quality data (\textbf{longer bar length} in Figure~\ref{fig:InsCL}).
We empirically demonstrate that replaying more data with high InsInfo helps to alleviate the forgetting issue.

The main contributions of this paper include:
(1) We propose InsCL, a novel replay-based CL paradigm for instruction tuning. 
InsCL allocates replay size based on task similarity, dynamically replaying high-quality data with high InsInfo.
(2) Experiments are conducted over 16 tasks with different training orders, demonstrating the effectiveness of InsCL. 
(3) We further analyze the forgetting phenomenon in continual instruction tuning.
Without replaying, we found that complex reasoning tasks suffer from a higher forgetting rate, where forgetting instances are mainly instruction-unrelated.

\begin{table}[t]
\small
\centering
\begin{tabular}{p{7cm}}
\toprule
\colorbox[RGB]{231,230,230}{\textbf{Instruction}}: In this task, you're given reviews from Amazon's products. Your task is to generate the Summary of the review. \\
\colorbox[RGB]{231,230,230}{\textbf{Input}}: Totally screwed up my system. Instructions terrible. Disk gives long list of files, had to determine what does what. Has already wasted 4 hours of my time. I gave up and pulled the thing. Don't buy this.
\newline
\colorbox[RGB]{231,230,230}{\textbf{Output}}: Terrible. Instructions are non-existent.\\
\bottomrule
\end{tabular}
\caption{A case of data template in instruction tuning.}
\label{tab:ins_case}
\end{table}

\section{Related Work}
\subsection{Instruction Tuning}
Recently, LLMs have demonstrated impressive performance across various NLP tasks.
After being unsupervised pre-trained on large-scale raw text, LLMs are further trained via instruction tuning to generate appropriate outputs based on the given input instructions~\cite{sanh2021multitask, mishra2021natural, chung2022scaling}. 
Prior works supervised fine-tuned (SFT) LLMs with datasets consisting of \{instruction, input, output\} pairs, as shown in Table~\ref{tab:ins_case}, and evaluated on another set of held-out tasks~\cite{wei2021finetuned, longpre2023flan}.
They demonstrate that the performance of unseen tasks can be improved with more tasks and templates.
To improve the diversity and complexity of instruction, a broad range of open-source instruction tuning datasets are proposed. 
Some are gathered through crowd-sourcing~\cite{DatabricksBlog2023DollyV2, zhou2023lima} while others are distilled from strong proprietary models~\cite{wang2022self, peng2023instruction, taori2023stanford}.

With the help of various low-cost methods of constructing high-quality templates, instruction datasets can expand easily over time as new tasks appear. 
When the data scale grows dynamically, we can easily obtain sufficient task-specific data.
Considering this, rather than evaluating zero-shot ability on held-out tasks, we are more concerned about adapting an instruction-tuned model to a new task without suffering from catastrophic forgetting.
In this work, we fine-tune LLMs in a continuous manner and analyze their performance on previous tasks, aiming to explore the forgetting issue in a changeable environment. 

\subsection{Traditional CL Methods}
CL aims to learn a sequence of tasks incrementally without forgetting the previously learned knowledge.
Early attempts in CL can be generally divided into three categories:
(1) \textbf{Consolidation-based methods} aim at protecting important parameters. As the representative of the regularization sub-family, EWC~\cite{kirkpatrick2017overcoming} constrains the loss based on parameter importance calculated by the fisher information matrix. 
Several works distill the model from the previous stage to keep relevant knowledge~\cite{zhang2020class, monaikul2021continual, liu2021lifelong, qin2021lfpt5}.
(2) \textbf{Architecture-based methods} add task-specific parameters to the base model for each task~\cite{rusu2016progressive, gu2020transformer, madotto2020continual}. By separating trainable parameters, the model can mitigate the impact on old tasks when updating parameters.
However, the model scale grows linearly when tasks increase, bringing inevitable memory costs.
(3) \textbf{Replay-based methods} store a small subset of previous training examples and replay when the new task comes. \citet{sun2019lamol, zhang2022continual} leverage language models to generate pseudo-examples for previous tasks, but the quality of examples cannot be guaranteed~\cite{ke2021achieving}.

Despite the success of traditional CL methods, their backbones are relatively small in scale, such as BERT~\cite{devlin2018bert} and RoBERTa~\cite{liu2019roberta}. 
Under LLMs' full fine-tuning scenarios, consolidation-based and architecture-based methods will bring additional parameter storage and training costs. 
Considering the heavy burden on computing time and GPU memory, replay-based CL methods are the simplest and most widely used in tuning LLMs as data-efficient methods that do not change the model structure.

\subsection{CL for LLMs instruction tuning}
Due to the scaling laws for neural language models, LLMs emerge with capabilities when the scale increases.
They can be better adapted to various downstream tasks through instruction tuning, offering immense practical value in real-world applications.
The exploration of CL for LLMs is still in its early stages.
Continual-T0~\cite{scialom2022fine} first fine-tuned LLMs with instructions in an incremental manner, claiming that well-pre-trained models can be continual learners by randomly replaying several previous examples.
Several works~\cite{song2023conpet, wang2023orthogonal} focus on CL methods with parameter-efficient tuning~\cite{hu2021lora}, largely alleviating the forgetting issue under limited training resources.
For full fine-tuning, replay-based methods were preliminarily investigated~\cite{yin2023dynosaur}, proving that replaying data based on diverse instructions can alleviate catastrophic forgetting and help better generalize to unseen tasks.
However, there is still a lack of detailed analysis of replay strategies.

In this work, we focus on the appropriate replay-based method for LLMs' full fine-tuning with instructions.
Considering that instructions naturally provide high-quality task-related descriptions, it is necessary to fully utilize instruction information to customize a replay strategy for instruction tuning.

\section{Method}
Continual Learning of LLMs focuses on adapting an instruction-tuned model to handle a sequence of tasks in a specific application scenario.
This approach accounts for consistently emerging materials while processing the tasks simultaneously.
%
We define $n$ tasks to be learned as a sequence $D = \{ D_1, \ldots, D_n\}$.
When LLMs are tuned with $i$-th task, we form a replay dataset $R_j^\alpha$ by sampling examples from $D_j$, where $j\in[1, i-1]$.
Formally, the training data augmented with replay data is defined as:

$$
D_i^\alpha = D_i \cup \sum_{j=1}^{i-1} R_j^\alpha
$$
where $\alpha$ is the replay hyper-parameter, controlling the sampling quantity from previous tasks.

\subsection{Dynamic Replay}
\label{sec:dynamic_replay}
Prior works optimize CL methods based on the similarity between previous tasks and the current one~\cite{mi2020continual, xu2023continual, gogoulou2023study}.
As the similarity increases, it becomes easier to retain knowledge from previous tasks.
Inspired by this, we propose a dynamic replay strategy based on task similarity, replaying more data from previous tasks with large differences.

The concept of task similarity is at the core of various machine learning paradigms, such as domain adaptation and meta-learning.
Optimal Transport~\cite{alvarez2020geometric, torres2021survey} offers a way to calculate the least amount of cost for transferring between different distribution pairs. 
As the representative of the Optimal Transport framework, Wasserstein Distance~\cite{chen2022inferential, liu2022wasserstein} provides a metric for calculating the similarity between two dataset distributions. 
The definition of Wasserstein Distance is as follows:
$$
W\left(\mu_A, \mu_B\right)=\inf_{\pi}\left( \int_{\mathbb{R}} d(x_A, x_B) d \pi(x_A, x_B)\right)
$$
where $\pi\in\prod\left(\mu_{A}, \mu_{B}\right)$ is meant to be the set of all joint probabilities that exhibit $\mu_{A}$ and $\mu_{B}$ as marginal distributions. 
The $d$ denotes a metric for calculating the cost matrix, and here we define it as the cosine distance.
For instruction tuning, NLP tasks can be described via natural language instructions.
We consider the instruction embeddings for a task pair as $x_A$ and $x_B$, and calculate the proportion of instructions for each task as a probability distribution.
Consequently, we measure task similarity by calculating their Wasserstein Distance.
When LLMs are fine-tuned on the current task $D_i$, the amount of dynamic replay data for the $j$-th previous task is defined as:
$$
\alpha_j^* = \frac{W_{j,i}}{\sum_{k=1}^{i-1} W_{k,i}} \times \alpha ,\quad j\in[1, i-1]
$$ 
where $W_{j,i}$ denotes the Wasserstein Distance between $D_j$ and $D_i$.
We dynamically allocate the amount of previous data to replay according to its similarity with the current task. 
With the help of dynamic replay, LLMs selectively recall the corresponding knowledge.

\subsection{Instruction Information Metric}
It has been proven that a small amount of high-quality data can achieve a promising performance, demonstrating the rationality of careful data selection~\cite{de2019episodic, wang2020efficient, ke2022continual, zhou2023lima}.
Inspired by this, we propose an Instruction Information Metric (InsInfo) to guide the sampling process, collecting high-quality replay data for continual instruction tuning.


Considering \textbf{complex} and \textbf{diverse} instructions induce impressive performance, a more comprehensive analysis of multiple intentions embedded within instructions is necessary.
High-performing open-source LLMs demonstrate the ability to annotate queries with tag entities, and the precision and consistency are proven through manual annotation~\cite{lu2023instag}.
Consequently, we employ GPT-4~\cite{openai2023gpt4} as an intention tagger and clean the raw tags, representing instructions at a fine-grained entity level.
The detailed process of obtaining normalized tags is shown in Appendix~\ref{sec:InsTag}.
After obtaining fine-grained annotations for instructions, we utilize the number and frequency of tags as quantifiable indicators of diversity and complexity.
Motivated by Inverse Document Frequency (IDF), one of the most useful and widely used concepts in information retrieval~\cite{gupta2022tidf, tayal2023automatic}, we proposed InsInfo as follows to quantify instruction information:
$$
{\rm InsInfo} = \sum_{t=1}^{T} log \frac{N}{f_t}
$$
where $N$ denotes the total amount of previous instructions.
When tasks come into a stream, we store all previous instructions in memory.
For each instruction, $T$ denotes the number of tags, and $f_t$ denotes the frequency of the $t$-th tag among the instruction pool.
Hence, instruction gets a large InsInfo when the number of individual tags increases, quantifying complexity and diversity interpretably.
As shown in Algorithm~\ref{alg:InsInfo}, we follow the InsInfo-guided sampling strategy to obtain the replay data.
Moreover, the strategy can be combined with dynamic replay by modifying $\alpha$ to $\alpha_j^*$, as claimed in Section~\ref{sec:dynamic_replay}, which forms our InsCL finally.

\begin{algorithm}[t]
\caption{InsInfo-guided sampling}\label{alg:InsInfo}
\KwData{Dataset $D_j$, Instruction Pool $I_{i}$, Replay Number $\alpha$ }
\KwResult{Replay dataset $R^{\alpha}_{j}$}
Initialize Empty $R^{\alpha}_{j}$ and InsInfo List $S_j$\;
Extract task $j$ instruction set $I_j$ from $I_i$\;
\For{Query $I_{j,k} \in I_j$}
{
$s_{j,k}$ $\leftarrow$ calculate InsInfo for $I_{j,k}$ \;
$S_j\leftarrow S_j \cup s_{j,k}$\;
}
\For{$k=1$ to $|I_j|$}
{
$\beta$ $\leftarrow$ $\frac{s_{j,k}}{sum \left( S_{j} \right) } \times \alpha$ \;
$D_{j,k}$ $\leftarrow$ \{data in $D_j$ with $I_{j,k}$\} \;
$R^{\alpha}_{j} \leftarrow$ sample $\beta$ data from $D_{j,k}$ \;
}
\textbf{return} $R^{\alpha}_{j}$
\end{algorithm}

\section{Experimental Setup}

\begin{figure}[t]
   \centering
   \includegraphics[width=0.8\linewidth]{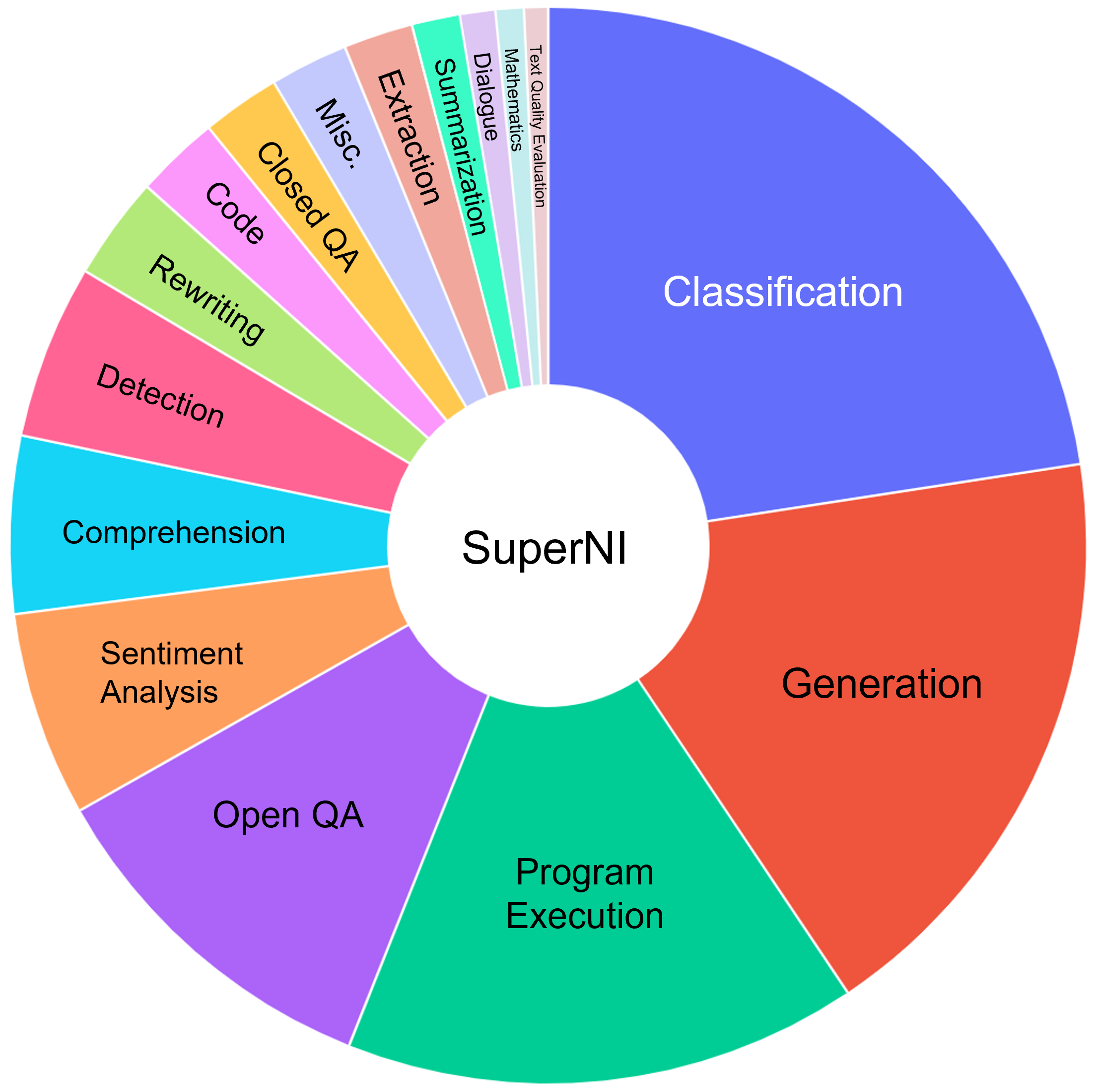}
   \caption{We obtain 16 categories by integrating English tasks in the SuperNI dataset. And we conduct further experiments based on 16 reallocated tasks.}
   \label{fig:category}
\end{figure}

\textbf{Data Collection.}\quad
To facilitate our research, we mainly utilize the SuperNI dataset~\cite{wang2022super}, a comprehensive benchmark focusing on specific NLP tasks distilled from real-world demands.
SuperNI is annotated by NLP practitioners from GitHub and NLP courses, ensuring that each instance is coupled with respective natural language instructions.
At the most comprehensive level, we integrate 765 English tasks from SuperNI into 16 categories, as shown in Figure~\ref{fig:category}.
And we demonstrate details of the data composition in Appendix~\ref{sec:Data_Composition}.
Following the setting of prior CL studies~\cite{scialom2022fine, yin2023dynosaur}, we randomly hold out $20\%$ instances on each task to evaluate LLMs on different training stages.

\noindent\textbf{Model and Training Details.}\quad
Our work is most related to the continual instruction tuning setting as Continual-T0~\cite{scialom2022fine}.
We conduct our task-incremental experiments with the popular LLaMA-7B~\cite{touvron2023llama}, training each task for 2 epochs with a batch size of 64.
We use the Adam optimizer~\cite{kingma2014adam} with a learning rate of 2e-5 and utilize the standard language modeling objective:
$$
\mathcal{L}=-\frac{1}{|y|} \sum_{i=1}^{|y|} \log p_{\theta}\left(y_{i} \mid x, y_{<i}\right)
$$
where $x$ denotes the combination of instruction and input, and $y$ denotes the corresponding output.

\noindent\textbf{Evaluate Forgetting.}\quad
Following the evaluation metric proposed by~\citet{scialom2022fine}, we leverage Relative Gain to focus on the forgetting issue.
We train expert LLM on each single task only and test with their respective holdout data, taking the results as upper bounds~\cite{jang2023exploring}.
The Relative Gain in stage $i$ can be defined as:
$$
{\rm \scalebox{0.85}{Relative\ Gain}}^i = \frac{1}{i-1} \sum_{j=1}^{i-1} \frac{R_j^i}{{\rm \scalebox{0.8}{upper\ bound}}_j} \times 100\%.
$$
Here we utilize Rouge-L~\cite{lin2004rouge} to calculate $R_j^i$ and the upper bound.

\begin{figure*}[t]
   \centering
   \includegraphics[width=0.97\linewidth]{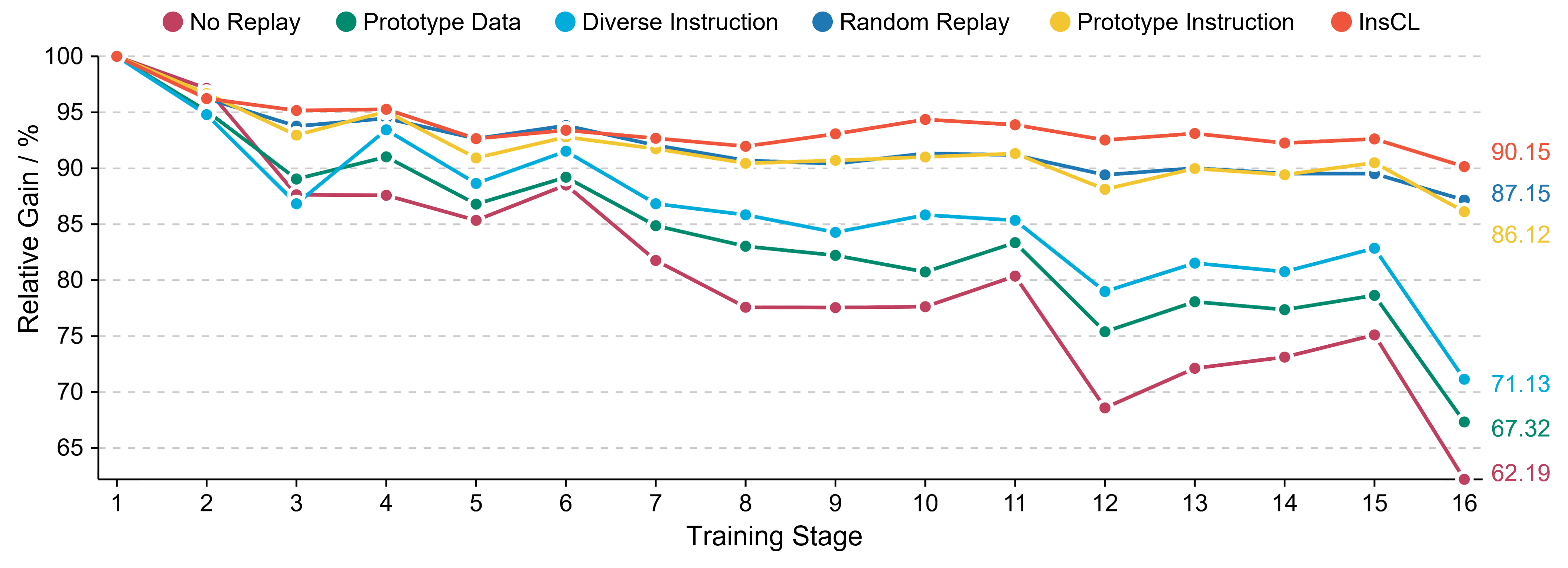}
   \caption{Progressive Relative Gain results for LLaMA-7B in continual instruction tuning. 
   We set Relative Gain to 100 for training on the first task, denoting the initial performance without forgetting.
   When it comes to stage $i$, we plot the average score of corresponding Relative Gain with three different training orders.
   The closer the Relative Gain is to 100, the better to alleviate catastrophic forgetting and preserve knowledge.}
   \label{fig:rouge}
\end{figure*}

\section{Experiments}

We leverage LLaMA-7B to calculate sentence embeddings and compare our InsCL with the following strategies:
\begin{itemize}
\setlength{\topsep}{0pt}
\setlength{\itemsep}{0pt}
\setlength{\parsep}{0pt}
\setlength{\parskip}{0pt}
\item \textbf{No Replay}: 
Train LLMs incrementally without any replay data.

\item \textbf{Random Replay}:
Sample $\alpha$ instances randomly from each previous task as the replay setting in Continual-T0.

\item \textbf{Prototype Data}: 
To collect the most representative data, we cluster the training data embedding space with k-means~\cite{wang2021mell}. For each previous task, we set the cluster number as the amount of instructions. 
We sort the data in descending order according to cosine distance from the corresponding center and take the top-$\alpha$ as replay data.

\item \textbf{Prototype Instruction}: 
We cluster instructions on previous tasks with the optimal silhouette coefficient~\cite{dinh2019estimating}, taking the closest instructions to their respective centers as the most representative.
We randomly select $\alpha$ data with prototypical instructions.

\item \textbf{Diverse Instruction}: 
Following the optimal replay strategy proposed by~\citet{yin2023dynosaur}, we replay data with instructions diverging most from the current task instructions.
By computing the cosine similarity matrix with the current instruction embedding, we take the most diverse instruction with the least column sum and replay $\alpha$ corresponding data for each previous task.
\end{itemize}

For fairness of comparison among different methods, we note $M_i = (i-1) \times \alpha$ as the total amount of replay data when the task sequence comes to stage $i$. 
Here we set $\alpha$ to 200.

\begin{table*}[t]
\setlength{\tabcolsep}{14pt}
\centering
\resizebox{1.95\columnwidth}{!}
{
\begin{tabular}{lcccccc}
\bottomrule
 &\multicolumn{2}{c}{\textbf{Reverse}}& \multicolumn{2}{c}{\textbf{Random}} & \multicolumn{2}{c}{\textbf{Curriculum}}\\
\hline
\textbf{Method}& \multicolumn{1}{c}{\textbf{AVG}} & \multicolumn{1}{c}{\textbf{STD}} & \multicolumn{1}{c}{\textbf{AVG}} & \multicolumn{1}{c}{\textbf{STD}} & \multicolumn{1}{c}{\textbf{AVG}} & \multicolumn{1}{c}{\textbf{STD}}\\
\hline 
No Replay &  73.83 & 182.87 & 81.07 & 121.9 & 87.63 & 51.30 \\
Random Replay & 87.96 & 18.85 & 92.90 & \underline{10.84} & \underline{95.18} & \underline{4.80} \\
Prototype Data & 78.07 & 92.71 & 83.51 & 93.71 & 90.07 & 29.79 \\
Prototype Instruction  & \underline{88.29} & \underline{15.73} & \underline{93.01} & 18.75 & 93.91 & 7.44 \\
Diverse Instruction  & 80.87 & 72.09 & 86.47 & 81.60 & 91.14 & 23.34 \\
\hline
InsCL & \textbf{90.50} & \textbf{9.32} & \textbf{94.43} & \textbf{7.62} & \textbf{96.20} & \textbf{2.81} \\
\bottomrule
\end{tabular}
}
\caption{\label{tab:order}
Results on different training orders.
AVG indicates average Relative Gain on 16 tasks, and STD indicates standard deviation ($\times$ e-4) on all the Relative Gain.
Reverse denotes a converse training order with Curriculum. 
A promising method is expected with a large AVG and a small STD, indicating good performance and high stability.
The best results are in bold, while the second-best are underlined.
}
\end{table*}

\subsection{Main Results}
We train LLaMA-7B on 16 tasks continuously with three different training orders.
For each continual instruction tuning stage, the average Relative Gain results are shown in Figure~\ref{fig:rouge}.
It can be observed that our InsCL is effective in mitigating forgetting, with a promising Relative Gain.
When all tasks have been trained, InsCL achieves performance gains of 3.0 Relative Gain compared with Random Replay, and 27.96 Relative Gain compared with No Replay.
InsCL sustainably maintains the performance on previous tasks over $90\%$, exhibiting high stability with a small fluctuation. 
Conversely, No Replay's Relative Gain shows a sharp decreasing trend as the task increases, accompanied by significant performance fluctuations.
After training the 8th task, No Replay's performance remains at less than $80\%$ and further drops to less than $65\%$ upon finishing final training.
No Replay setting severely suffers from catastrophic forgetting, demonstrating the necessity of replaying previous data.

Moreover, we further analyze other replay-based methods.
Despite being the optimal method in the previous work, Diverse Instruction underperforms when compared with Random Replay and Prototype Instruction.
For prototype-based methods, Prototype Instruction outperforms Prototype Data.
We find that clustering results of Prototype Data are significantly affected by instances with long instruction and short input, leading to practically identical embeddings for this subset.
The uneven distribution will cause a high semantic duplicate selection, which has been proven to lead to a negative impact~\cite{abbas2023semdedup}.
The data composed of instruction and input has a different structure from traditional SFT, resulting in several traditional replay-based methods not being directly applicable to instruction tuning.
This observation also demonstrates the rationality of designing instruction-based replay methods, proving the consistency of our InsCL.

\begin{figure*}[t]
   \centering
   \includegraphics[width=1.0\linewidth]{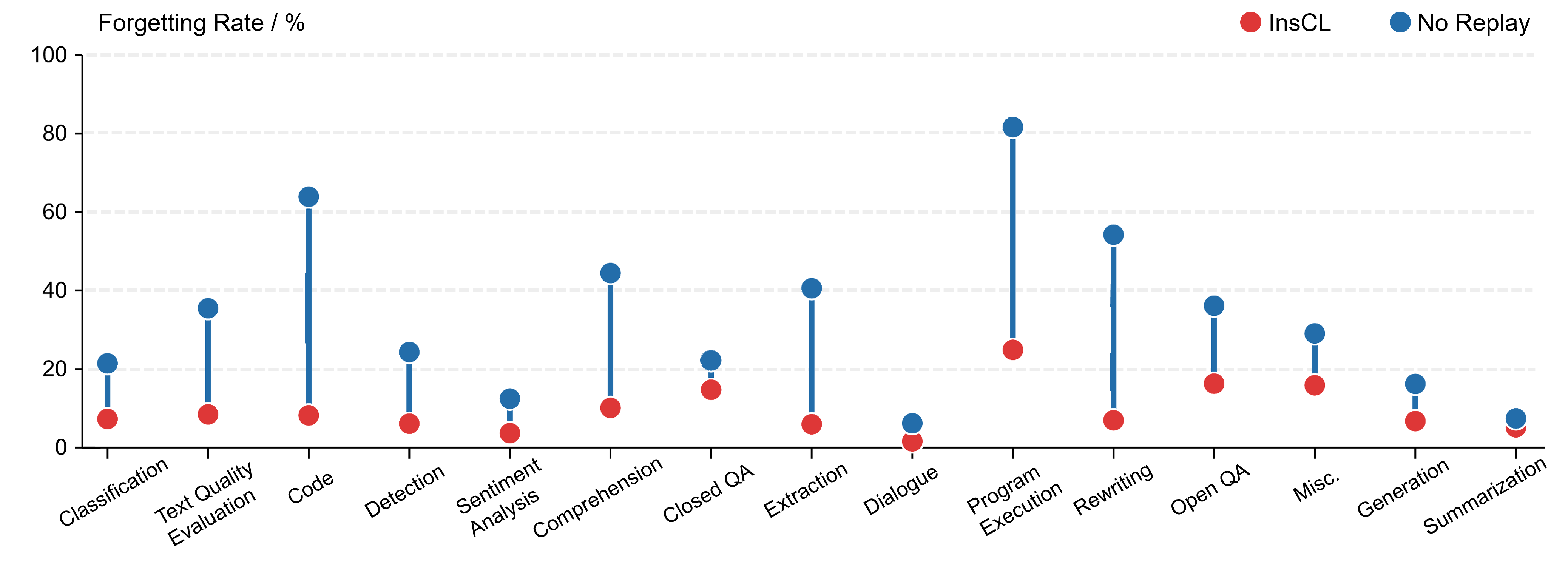}
   \caption{We analyze the forgetting rate based on Curriculum training order. 
   The results of all previous tasks are reported when training is finished on the last task.}
   \label{fig:FG}
\end{figure*}

\subsection{Training Order Analysis}
To explore the impact of training order and obtain universe conclusions, we conduct a detailed analysis of all settings based on different task sequences.
Inspired by Curriculum Learning~\cite{wang2021survey}, we train the model from easy task to hard task by sorting the upper bounds in descending order, as
\emph{
Classification $\rightarrow$
Text Quality Evaluation $\rightarrow$
Code $\rightarrow$
Detection $\rightarrow$
Sentiment Analysis $\rightarrow$
Comprehension $\rightarrow$
Closed QA $\rightarrow$
Extraction $\rightarrow$
Dialogue $\rightarrow$
Program Execution $\rightarrow$
Rewriting $\rightarrow$
Open QA $\rightarrow$
Misc. $\rightarrow$
Generation $\rightarrow$
Summarization$\rightarrow$
Mathematics.
}

As shown in Table~\ref{tab:order}, we report the average Relative Gain scores and the standard deviations on 16 tasks with different training orders.
When we utilize the "easy to hard" training strategy, Curriculum outperforms other orders in all CL methods.
Under the No Replay setting, Curriculum achieves performance gains of $13.80$ average Relative Gain compared with Reverse and $6.56$ compared with Random.
Training tasks in Curriculum order demonstrates a more stable performance with a small standard deviation.
Moreover, with our InsCL, Curriculum achieves performance gains of $5.70$ average Relative Gain compared with Reverse and $1.77$ compared with Random.
It can be observed that InsCL alleviates the impact of different training orders, outperforming all methods with a high Relative Gain and stability.

\begin{table}[t]
\setlength{\tabcolsep}{9pt}
\centering
\resizebox{0.97\columnwidth}{!}
{
\begin{tabular}{lcc}
\bottomrule
\textbf{Method}& \textbf{AVG} &\textbf{STD}\\
\hline
No Replay & 80.84 & 118.69 \\
\hline
Random Replay  & 92.01 & 11.50 \\
 \hspace{4mm} + Dynamic (Uniform)  & 93.14 & 8.67 \\
 \hspace{4mm} + Dynamic (Real)  & 93.25 & \underline{8.57} \\
 \hspace{4mm} + InsInfo & \underline{93.52} & 17.90 \\
\hline 

InsCL & \textbf{93.71} & \textbf{6.58} \\

\bottomrule
\end{tabular}
}
\caption{\label{tab:ablation}
Average results on three training orders.
AVG indicates average Relative Gain, and STD indicates standard deviation ($\times$ e-4) on all the Relative Gain.
The best results are in bold, while the second-best are underlined.
}
\end{table}

\subsection{Ablation Study}
To investigate the effectiveness of each component in InsCL, we further apply our dynamic replay and InsInfo-guided sampling based on the Random Replay.
Dynamic replay is determined by task similarity, calculated via Wasserstein distance.
If the real distribution of instructions cannot be obtained, the uniform distribution assumption is generally used to obtain the Wasserstein distance.
We evaluate the performance with average Relative Gain scores and standard deviations on all training stages.

The average results over three different training orders are reported in Table~\ref{tab:ablation}.
It can be inferred that dynamic replay and InsInfo-guided sampling are both beneficial to mitigating catastrophic forgetting.
InsInfo-guided sampling brings greater improvement in Relative Gain, effectively improving Relative Gain but lacking in stability.
Instead, dynamic replay greatly reduces the standard deviation of Relative Gain thus improving stability.
And dynamic replay with real distribution brings better performance compared with the uniform distribution assumption.
When we utilize InsCL combined with dynamic replay and InsInfo-guided sampling, it achieves the best performance and strongest stability. 
Compared with Random Replay, InsCL delivers an improved average Relative Gain of 1.71 and a reduced standard deviation of 4.92.
Furthermore, when compared with No Replay, InsCL achieves an improved average Relative Gain of 12.87 and a dramatic reduction of the standard deviation.
The results prove the effectiveness of each component and demonstrate that InsCL leverages the strengths of each.

\begin{figure*}[t]
   \centering
   \includegraphics[width=1.0\linewidth]{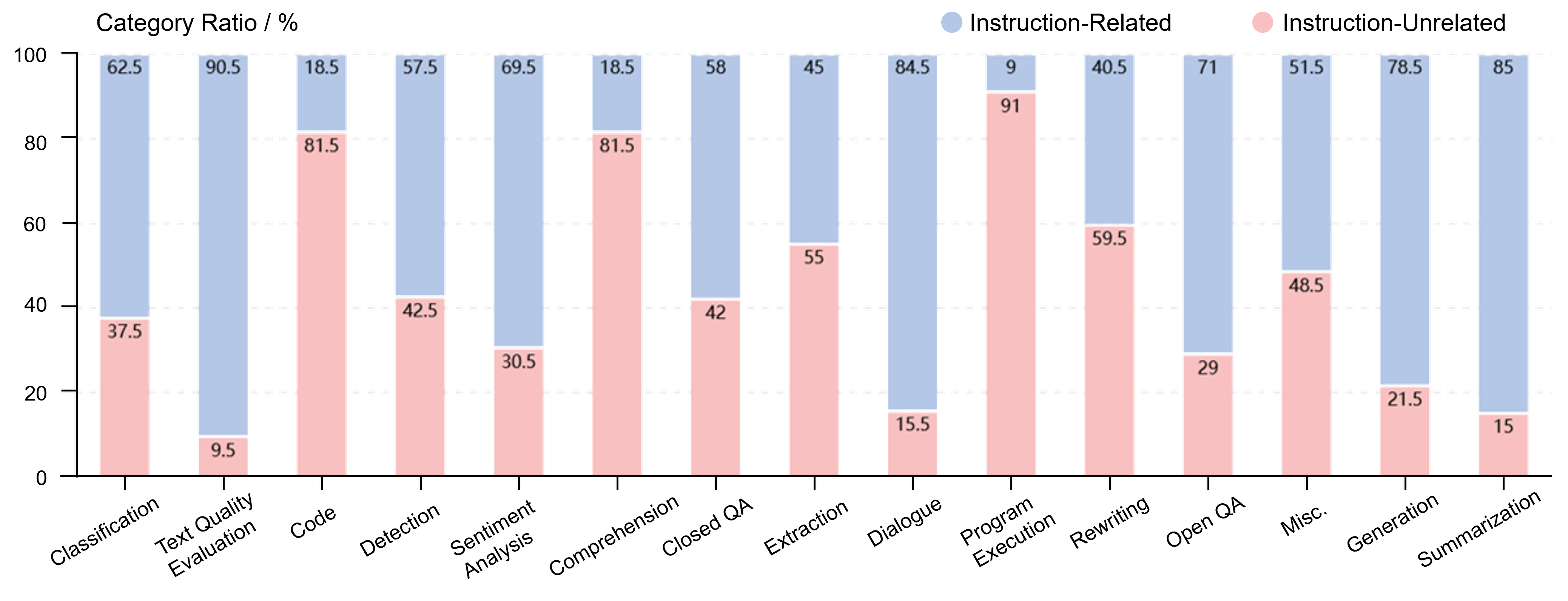}
   \caption{The analysis of forgetting category. 
   We divide forgetting instances into Instruction-Related and Instruction Unrelated. 
   After training on Curriculum order, the ratios of two categories in previous tasks are reported.}
   \label{fig:FG_category}
\end{figure*}

\subsection{Forgetting Analysis}
\textbf{Forgetting Rate.}\quad 
For a further catastrophic forgetting analysis, several methods~\cite{kemker2018measuring, luo2023empirical} quantify the forgetting issue by evaluating performance decrease as training incrementally.
Consequently, we propose a forgetting rate defined as:
$$
FG_i = \frac{R_i^{*} - R_i^{-1}}{R_i^{*}} \times 100\%
$$
where $R_i^{*}$ is the initial Rouge-L of task $i$ after training on the corresponding task, and $R_i^{-1}$ is the final Rouge-L of task $i$ in the last training stage.

We evaluate the forgetting rate with Curriculum training order and report the results of No Replay and InsCL in Figure~\ref{fig:FG}. 
It can be inferred that there is no inevitable relationship between task order and forgetting rate.
For tasks that require complex reasoning, Program Execution and Code severely suffer from forgetting with the No Replay setting.
Additionally, a large training data scale does not necessarily lead to a small forgetting rate.
For example, Classification and Generation are the top-2 tasks with large training data and exhibit smaller forgetting rates, while Program Execution with the third largest dataset suffers from the largest forgetting rate.
With our InsCL, the forgetting rates of almost all tasks are below 20\%, which means that most of the previous knowledge is preserved.

\noindent\textbf{Forgetting Category.}\quad
When all the tasks have been trained under the No Replay setting, we collect previous tasks' instances with a decreased Rouge-L, called forgetting instances. 
We randomly sampled 200 forgetting instances from each previous task, manually analyzing the forgetting category for a detailed conclusion.
We divide forgetting instances into two categories based on the instruction's following ability: 
(1) Instruction-Related: 
The output is relevant to the instruction, according to the space defined by the instruction. 
This category indicates LLMs do not forget the corresponding instruction following ability.
(2) Instruction-Unrelated: 
The output is unrelated to the instruction.
We demonstrate representative cases and respective explanations in Appendix~\ref{sec:FG_category_detail}.

Figure~\ref{fig:FG_category} reports category ratios in the curriculum training order.
The forgotten instances of most tasks are mainly Instruction-Related, while the forgetting instances in 5 tasks are mainly Instruction-Unrelated.
Additionally, more than 80\% of forgetting instances in Program Execution, Code, and Comprehension tasks are Instruction-Unrelated.
It can be inferred that failure to understand instructions mainly leads to the performance decline of complex reasoning tasks.

\section{Conclusions}

In this paper, we mainly discuss the efficient adaptation of LLMs to continual downstream tasks with instructions.
Replay-based CL methods do not require additional modifications to LLMs and fully utilize previous data, mitigating catastrophic forgetting effectively.
We proposed InsCL, an effective data-efficient method to mitigate catastrophic forgetting for LLMs instruction tuning.
InsCL is a model-agnostic and training-free method, indicating strong transferability.
Different from existing replay-based methods, we fully utilize instructions as representative task descriptions to design the replay strategy.
InsCL leverages instruction embeddings and distributions to calculate Wasserstein distance for task similarity, adjusting the replay ratio dynamically.
Then, with our InsInfo-guided sampling, InsCL selects more high-quality data with complex and diverse instructions.
We conduct extensive experiments over 16 tasks with different training orders, observing consistent performance improvements of InsCL.
Additionally, we further analyze the forgetting rate and forgetting category, aiming to provide a guideline for future work.

\section{Limitations}
The promising performance demonstrated by InsCL is dependent on high-quality instructions.
Instead, fuzzy instructions can affect the calculation of task similarity and the InsInfo-guided sampling, which may mislead our InsCL.
However, if the instruction-based dataset is unsatisfied, the performance of tuned LLMs will also be greatly affected.
Therefore, we tend to use our method after collecting high-quality instruction-based data to further mitigate catastrophic forgetting.

\section{Acknowledgments}
This work was supported by the Shenzhen Science and Technology under Grant JSGG202208311102\\
03007.

\bibliography{anthology, custom}

\begin{thebibliography}{69}
\expandafter\ifx\csname natexlab\endcsname\relax\def\natexlab#1{#1}\fi

\bibitem[{Abbas et~al.(2023)Abbas, Tirumala, Simig, Ganguli, and Morcos}]{abbas2023semdedup}
Amro Abbas, Kushal Tirumala, D{\'a}niel Simig, Surya Ganguli, and Ari~S Morcos. 2023.
\newblock Semdedup: Data-efficient learning at web-scale through semantic deduplication.
\newblock \emph{arXiv preprint arXiv:2303.09540}.

\bibitem[{Alvarez-Melis and Fusi(2020)}]{alvarez2020geometric}
David Alvarez-Melis and Nicolo Fusi. 2020.
\newblock Geometric dataset distances via optimal transport.
\newblock \emph{Advances in Neural Information Processing Systems}, 33:21428--21439.

\bibitem[{Bird et~al.(2009)Bird, Klein, and Loper}]{bird2009natural}
Steven Bird, Ewan Klein, and Edward Loper. 2009.
\newblock \emph{Natural language processing with Python: analyzing text with the natural language toolkit}.
\newblock " O'Reilly Media, Inc.".

\bibitem[{Brown et~al.(2020)Brown, Mann, Ryder, Subbiah, Kaplan, Dhariwal, Neelakantan, Shyam, Sastry, Askell et~al.}]{brown2020language}
Tom Brown, Benjamin Mann, Nick Ryder, Melanie Subbiah, Jared~D Kaplan, Prafulla Dhariwal, Arvind Neelakantan, Pranav Shyam, Girish Sastry, Amanda Askell, et~al. 2020.
\newblock Language models are few-shot learners.
\newblock \emph{Advances in neural information processing systems}, 33:1877--1901.

\bibitem[{Chen et~al.(2022)Chen, Gao, and Wang}]{chen2022inferential}
Yao Chen, Qingyi Gao, and Xiao Wang. 2022.
\newblock Inferential wasserstein generative adversarial networks.
\newblock \emph{Journal of the Royal Statistical Society Series B: Statistical Methodology}, 84(1):83--113.

\bibitem[{Chung et~al.(2022)Chung, Hou, Longpre, Zoph, Tay, Fedus, Li, Wang, Dehghani, Brahma et~al.}]{chung2022scaling}
Hyung~Won Chung, Le~Hou, Shayne Longpre, Barret Zoph, Yi~Tay, William Fedus, Yunxuan Li, Xuezhi Wang, Mostafa Dehghani, Siddhartha Brahma, et~al. 2022.
\newblock Scaling instruction-finetuned language models.
\newblock \emph{arXiv preprint arXiv:2210.11416}.

\bibitem[{Conover et~al.(2023)Conover, Hayes, Mathur, Xie, Wan, Shah, Ghodsi, Wendell, Zaharia, and Xin}]{DatabricksBlog2023DollyV2}
Mike Conover, Matt Hayes, Ankit Mathur, Jianwei Xie, Jun Wan, Sam Shah, Ali Ghodsi, Patrick Wendell, Matei Zaharia, and Reynold Xin. 2023.
\newblock \href {https://www.databricks.com/blog/2023/04/12/dolly-first-open-commercially-viable-instruction-tuned-llm} {Free dolly: Introducing the world's first truly open instruction-tuned llm}.

\bibitem[{Cuturi(2013)}]{cuturi2013sinkhorn}
Marco Cuturi. 2013.
\newblock Sinkhorn distances: Lightspeed computation of optimal transport.
\newblock \emph{Advances in neural information processing systems}, 26.

\bibitem[{Daruna et~al.(2021)Daruna, Gupta, Sridharan, and Chernova}]{daruna2021continual}
Angel Daruna, Mehul Gupta, Mohan Sridharan, and Sonia Chernova. 2021.
\newblock Continual learning of knowledge graph embeddings.
\newblock \emph{IEEE Robotics and Automation Letters}, 6(2):1128--1135.

\bibitem[{de~Masson~D'Autume et~al.(2019)de~Masson~D'Autume, Ruder, Kong, and Yogatama}]{de2019episodic}
Cyprien de~Masson~D'Autume, Sebastian Ruder, Lingpeng Kong, and Dani Yogatama. 2019.
\newblock Episodic memory in lifelong language learning.
\newblock \emph{Advances in Neural Information Processing Systems}, 32.

\bibitem[{Devlin et~al.(2018)Devlin, Chang, Lee, and Toutanova}]{devlin2018bert}
Jacob Devlin, Ming-Wei Chang, Kenton Lee, and Kristina Toutanova. 2018.
\newblock Bert: Pre-training of deep bidirectional transformers for language understanding.
\newblock \emph{arXiv preprint arXiv:1810.04805}.

\bibitem[{Ding et~al.(2023)Ding, Chen, Xu, Qin, Zheng, Hu, Liu, Sun, and Zhou}]{ding2023enhancing}
Ning Ding, Yulin Chen, Bokai Xu, Yujia Qin, Zhi Zheng, Shengding Hu, Zhiyuan Liu, Maosong Sun, and Bowen Zhou. 2023.
\newblock Enhancing chat language models by scaling high-quality instructional conversations.
\newblock \emph{arXiv preprint arXiv:2305.14233}.

\bibitem[{Dinh et~al.(2019)Dinh, Fujinami, and Huynh}]{dinh2019estimating}
Duy-Tai Dinh, Tsutomu Fujinami, and Van-Nam Huynh. 2019.
\newblock Estimating the optimal number of clusters in categorical data clustering by silhouette coefficient.
\newblock In \emph{Knowledge and Systems Sciences: 20th International Symposium, KSS 2019, Da Nang, Vietnam, November 29--December 1, 2019, Proceedings 20}, pages 1--17. Springer.

\bibitem[{Gogoulou et~al.(2023)Gogoulou, Lesort, Boman, and Nivre}]{gogoulou2023study}
Evangelia Gogoulou, Timoth{\'e}e Lesort, Magnus Boman, and Joakim Nivre. 2023.
\newblock A study of continual learning under language shift.
\newblock \emph{arXiv preprint arXiv:2311.01200}.

\bibitem[{Goodfellow et~al.(2013)Goodfellow, Mirza, Xiao, Courville, and Bengio}]{goodfellow2013empirical}
Ian~J Goodfellow, Mehdi Mirza, Da~Xiao, Aaron Courville, and Yoshua Bengio. 2013.
\newblock An empirical investigation of catastrophic forgetting in gradient-based neural networks.
\newblock \emph{arXiv preprint arXiv:1312.6211}.

\bibitem[{Gu et~al.(2020)Gu, Liu, Yu, Li, Chen, and Han}]{gu2020transformer}
Xiaotao Gu, Liyuan Liu, Hongkun Yu, Jing Li, Chen Chen, and Jiawei Han. 2020.
\newblock On the transformer growth for progressive bert training.
\newblock \emph{arXiv preprint arXiv:2010.12562}.

\bibitem[{Gupta et~al.(2022)Gupta, Mittal, Tiwari, Agarwal, and Singh}]{gupta2022tidf}
Ishu Gupta, Sloni Mittal, Ankit Tiwari, Priya Agarwal, and Ashutosh~Kumar Singh. 2022.
\newblock Tidf-dlpm: Term and inverse document frequency based data leakage prevention model.
\newblock \emph{arXiv preprint arXiv:2203.05367}.

\bibitem[{Hahsler et~al.(2019)Hahsler, Piekenbrock, and Doran}]{hahsler2019dbscan}
Michael Hahsler, Matthew Piekenbrock, and Derek Doran. 2019.
\newblock dbscan: Fast density-based clustering with r.
\newblock \emph{Journal of Statistical Software}, 91:1--30.

\bibitem[{Hu et~al.(2021)Hu, Shen, Wallis, Allen-Zhu, Li, Wang, Wang, and Chen}]{hu2021lora}
Edward~J Hu, Yelong Shen, Phillip Wallis, Zeyuan Allen-Zhu, Yuanzhi Li, Shean Wang, Lu~Wang, and Weizhu Chen. 2021.
\newblock Lora: Low-rank adaptation of large language models.
\newblock \emph{arXiv preprint arXiv:2106.09685}.

\bibitem[{Jang et~al.(2023)Jang, Kim, Ye, Kim, Logeswaran, Lee, Lee, and Seo}]{jang2023exploring}
Joel Jang, Seungone Kim, Seonghyeon Ye, Doyoung Kim, Lajanugen Logeswaran, Moontae Lee, Kyungjae Lee, and Minjoon Seo. 2023.
\newblock Exploring the benefits of training expert language models over instruction tuning.
\newblock \emph{arXiv preprint arXiv:2302.03202}.

\bibitem[{Jin et~al.(2021)Jin, Zhang, Zhu, Xiao, Li, Wei, Arnold, and Ren}]{jin2021lifelong}
Xisen Jin, Dejiao Zhang, Henghui Zhu, Wei Xiao, Shang-Wen Li, Xiaokai Wei, Andrew Arnold, and Xiang Ren. 2021.
\newblock Lifelong pretraining: Continually adapting language models to emerging corpora.
\newblock \emph{arXiv preprint arXiv:2110.08534}.

\bibitem[{Ke and Liu(2022)}]{ke2022continual}
Zixuan Ke and Bing Liu. 2022.
\newblock Continual learning of natural language processing tasks: A survey.
\newblock \emph{arXiv preprint arXiv:2211.12701}.

\bibitem[{Ke et~al.(2021)Ke, Liu, Ma, Xu, and Shu}]{ke2021achieving}
Zixuan Ke, Bing Liu, Nianzu Ma, Hu~Xu, and Lei Shu. 2021.
\newblock Achieving forgetting prevention and knowledge transfer in continual learning.
\newblock \emph{Advances in Neural Information Processing Systems}, 34:22443--22456.

\bibitem[{Kemker et~al.(2018)Kemker, McClure, Abitino, Hayes, and Kanan}]{kemker2018measuring}
Ronald Kemker, Marc McClure, Angelina Abitino, Tyler Hayes, and Christopher Kanan. 2018.
\newblock Measuring catastrophic forgetting in neural networks.
\newblock In \emph{Proceedings of the AAAI conference on artificial intelligence}.

\bibitem[{Kingma and Ba(2014)}]{kingma2014adam}
Diederik~P Kingma and Jimmy Ba. 2014.
\newblock Adam: A method for stochastic optimization.
\newblock \emph{arXiv preprint arXiv:1412.6980}.

\bibitem[{Kirkpatrick et~al.(2017)Kirkpatrick, Pascanu, Rabinowitz, Veness, Desjardins, Rusu, Milan, Quan, Ramalho, Grabska-Barwinska et~al.}]{kirkpatrick2017overcoming}
James Kirkpatrick, Razvan Pascanu, Neil Rabinowitz, Joel Veness, Guillaume Desjardins, Andrei~A Rusu, Kieran Milan, John Quan, Tiago Ramalho, Agnieszka Grabska-Barwinska, et~al. 2017.
\newblock Overcoming catastrophic forgetting in neural networks.
\newblock \emph{Proceedings of the national academy of sciences}, 114(13):3521--3526.

\bibitem[{Lin(2004)}]{lin2004rouge}
Chin-Yew Lin. 2004.
\newblock Rouge: A package for automatic evaluation of summaries.
\newblock In \emph{Text summarization branches out}, pages 74--81.

\bibitem[{Liu et~al.(2021)Liu, Yu, He, Liu, and Zhao}]{liu2021lifelong}
Qingbin Liu, Xiaoyan Yu, Shizhu He, Kang Liu, and Jun Zhao. 2021.
\newblock Lifelong intent detection via multi-strategy rebalancing.
\newblock \emph{arXiv preprint arXiv:2108.04445}.

\bibitem[{Liu et~al.(2022)Liu, Bai, Lu, Soltoggio, and Kolouri}]{liu2022wasserstein}
Xinran Liu, Yikun Bai, Yuzhe Lu, Andrea Soltoggio, and Soheil Kolouri. 2022.
\newblock Wasserstein task embedding for measuring task similarities.
\newblock \emph{arXiv preprint arXiv:2208.11726}.

\bibitem[{Liu et~al.(2019)Liu, Ott, Goyal, Du, Joshi, Chen, Levy, Lewis, Zettlemoyer, and Stoyanov}]{liu2019roberta}
Yinhan Liu, Myle Ott, Naman Goyal, Jingfei Du, Mandar Joshi, Danqi Chen, Omer Levy, Mike Lewis, Luke Zettlemoyer, and Veselin Stoyanov. 2019.
\newblock Roberta: A robustly optimized bert pretraining approach.
\newblock \emph{arXiv preprint arXiv:1907.11692}.

\bibitem[{Longpre et~al.(2023)Longpre, Hou, Vu, Webson, Chung, Tay, Zhou, Le, Zoph, Wei et~al.}]{longpre2023flan}
Shayne Longpre, Le~Hou, Tu~Vu, Albert Webson, Hyung~Won Chung, Yi~Tay, Denny Zhou, Quoc~V Le, Barret Zoph, Jason Wei, et~al. 2023.
\newblock The flan collection: Designing data and methods for effective instruction tuning.
\newblock \emph{arXiv preprint arXiv:2301.13688}.

\bibitem[{Lu et~al.(2023)Lu, Yuan, Yuan, Lin, Lin, Tan, Zhou, and Zhou}]{lu2023instag}
Keming Lu, Hongyi Yuan, Zheng Yuan, Runji Lin, Junyang Lin, Chuanqi Tan, Chang Zhou, and Jingren Zhou. 2023.
\newblock \# instag: Instruction tagging for analyzing supervised fine-tuning of large language models.
\newblock \emph{arXiv e-prints}, pages arXiv--2308.

\bibitem[{Luo et~al.(2023)Luo, Yang, Meng, Li, Zhou, and Zhang}]{luo2023empirical}
Yun Luo, Zhen Yang, Fandong Meng, Yafu Li, Jie Zhou, and Yue Zhang. 2023.
\newblock An empirical study of catastrophic forgetting in large language models during continual fine-tuning.
\newblock \emph{arXiv preprint arXiv:2308.08747}.

\bibitem[{Madotto et~al.(2020)Madotto, Lin, Zhou, Moon, Crook, Liu, Yu, Cho, and Wang}]{madotto2020continual}
Andrea Madotto, Zhaojiang Lin, Zhenpeng Zhou, Seungwhan Moon, Paul Crook, Bing Liu, Zhou Yu, Eunjoon Cho, and Zhiguang Wang. 2020.
\newblock Continual learning in task-oriented dialogue systems.
\newblock \emph{arXiv preprint arXiv:2012.15504}.

\bibitem[{Mi et~al.(2020)Mi, Chen, Zhao, Huang, and Faltings}]{mi2020continual}
Fei Mi, Liangwei Chen, Mengjie Zhao, Minlie Huang, and Boi Faltings. 2020.
\newblock Continual learning for natural language generation in task-oriented dialog systems.
\newblock \emph{arXiv preprint arXiv:2010.00910}.

\bibitem[{Mishra et~al.(2021)Mishra, Khashabi, Baral, and Hajishirzi}]{mishra2021natural}
Swaroop Mishra, Daniel Khashabi, Chitta Baral, and Hannaneh Hajishirzi. 2021.
\newblock Natural instructions: Benchmarking generalization to new tasks from natural language instructions.
\newblock \emph{arXiv preprint arXiv:2104.08773}, pages 839--849.

\bibitem[{Monaikul et~al.(2021)Monaikul, Castellucci, Filice, and Rokhlenko}]{monaikul2021continual}
Natawut Monaikul, Giuseppe Castellucci, Simone Filice, and Oleg Rokhlenko. 2021.
\newblock Continual learning for named entity recognition.
\newblock In \emph{Proceedings of the AAAI Conference on Artificial Intelligence}.

\bibitem[{OpenAI(2023)}]{openai2023gpt4}
OpenAI. 2023.
\newblock \href {http://arxiv.org/abs/2303.08774} {Gpt-4 technical report}.

\bibitem[{Ouyang et~al.(2022)Ouyang, Wu, Jiang, Almeida, Wainwright, Mishkin, Zhang, Agarwal, Slama, Ray et~al.}]{ouyang2022training}
Long Ouyang, Jeffrey Wu, Xu~Jiang, Diogo Almeida, Carroll Wainwright, Pamela Mishkin, Chong Zhang, Sandhini Agarwal, Katarina Slama, Alex Ray, et~al. 2022.
\newblock Training language models to follow instructions with human feedback.
\newblock \emph{Advances in Neural Information Processing Systems}, 35:27730--27744.

\bibitem[{Peng et~al.(2023)Peng, Li, He, Galley, and Gao}]{peng2023instruction}
Baolin Peng, Chunyuan Li, Pengcheng He, Michel Galley, and Jianfeng Gao. 2023.
\newblock Instruction tuning with gpt-4.
\newblock \emph{arXiv preprint arXiv:2304.03277}.

\bibitem[{Qin and Joty(2021)}]{qin2021lfpt5}
Chengwei Qin and Shafiq Joty. 2021.
\newblock Lfpt5: A unified framework for lifelong few-shot language learning based on prompt tuning of t5.
\newblock \emph{arXiv preprint arXiv:2110.07298}.

\bibitem[{Qin et~al.(2022)Qin, Zhang, Lin, Liu, Li, Sun, and Zhou}]{qin2022elle}
Yujia Qin, Jiajie Zhang, Yankai Lin, Zhiyuan Liu, Peng Li, Maosong Sun, and Jie Zhou. 2022.
\newblock Elle: Efficient lifelong pre-training for emerging data.
\newblock \emph{arXiv preprint arXiv:2203.06311}.

\bibitem[{Rusu et~al.(2016)Rusu, Rabinowitz, Desjardins, Soyer, Kirkpatrick, Kavukcuoglu, Pascanu, and Hadsell}]{rusu2016progressive}
Andrei~A Rusu, Neil~C Rabinowitz, Guillaume Desjardins, Hubert Soyer, James Kirkpatrick, Koray Kavukcuoglu, Razvan Pascanu, and Raia Hadsell. 2016.
\newblock Progressive neural networks.
\newblock \emph{arXiv preprint arXiv:1606.04671}.

\bibitem[{Sanh et~al.(2021)Sanh, Webson, Raffel, Bach, Sutawika, Alyafeai, Chaffin, Stiegler, Scao, Raja et~al.}]{sanh2021multitask}
Victor Sanh, Albert Webson, Colin Raffel, Stephen~H Bach, Lintang Sutawika, Zaid Alyafeai, Antoine Chaffin, Arnaud Stiegler, Teven~Le Scao, Arun Raja, et~al. 2021.
\newblock Multitask prompted training enables zero-shot task generalization.
\newblock \emph{arXiv preprint arXiv:2110.08207}.

\bibitem[{Scialom et~al.(2022)Scialom, Chakrabarty, and Muresan}]{scialom2022fine}
Thomas Scialom, Tuhin Chakrabarty, and Smaranda Muresan. 2022.
\newblock Fine-tuned language models are continual learners.
\newblock In \emph{Proceedings of the 2022 Conference on Empirical Methods in Natural Language Processing}, pages 6107--6122.

\bibitem[{Shi et~al.(2023)Shi, Su, Yang, Yang, and Cai}]{shi2023specialist}
Chufan Shi, Yixuan Su, Cheng Yang, Yujiu Yang, and Deng Cai. 2023.
\newblock Specialist or generalist? instruction tuning for specific nlp tasks.
\newblock In \emph{Proceedings of the 2023 Conference on Empirical Methods in Natural Language Processing}, pages 15336--15348.

\bibitem[{Shi et~al.(2024)Shi, Yang, Cai, Zhang, Wang, Yang, and Lam}]{shi2024thorough}
Chufan Shi, Haoran Yang, Deng Cai, Zhisong Zhang, Yifan Wang, Yujiu Yang, and Wai Lam. 2024.
\newblock A thorough examination of decoding methods in the era of llms.
\newblock \emph{arXiv preprint arXiv:2402.06925}.

\bibitem[{Song et~al.(2023)Song, Han, Zeng, Li, Chen, Liu, Sun, and Yang}]{song2023conpet}
Chenyang Song, Xu~Han, Zheni Zeng, Kuai Li, Chen Chen, Zhiyuan Liu, Maosong Sun, and Tao Yang. 2023.
\newblock Conpet: Continual parameter-efficient tuning for large language models.
\newblock \emph{arXiv preprint arXiv:2309.14763}.

\bibitem[{Sun et~al.(2019)Sun, Ho, and Lee}]{sun2019lamol}
Fan-Keng Sun, Cheng-Hao Ho, and Hung-Yi Lee. 2019.
\newblock Lamol: Language modeling for lifelong language learning.
\newblock \emph{arXiv preprint arXiv:1909.03329}.

\bibitem[{Taori et~al.(2023)Taori, Gulrajani, Zhang, Dubois, Li, Guestrin, Liang, and Hashimoto}]{taori2023stanford}
Rohan Taori, Ishaan Gulrajani, Tianyi Zhang, Yann Dubois, Xuechen Li, Carlos Guestrin, Percy Liang, and Tatsunori~B Hashimoto. 2023.
\newblock Stanford alpaca: An instruction-following llama model.

\bibitem[{Tayal et~al.(2023)Tayal, Bajaj, Gore, Yadav, and Chouhan}]{tayal2023automatic}
Madhuri~A Tayal, Vanshika Bajaj, Ankita Gore, Preeti Yadav, and Vaishnavi Chouhan. 2023.
\newblock Automatic domain classification of text using machine learning.
\newblock In \emph{2023 International Conference on Communication, Circuits, and Systems (IC3S)}, pages 1--5. IEEE.

\bibitem[{Tirumala et~al.(2023)Tirumala, Simig, Aghajanyan, and Morcos}]{tirumala2023d4}
Kushal Tirumala, Daniel Simig, Armen Aghajanyan, and Ari~S Morcos. 2023.
\newblock D4: Improving llm pretraining via document de-duplication and diversification.
\newblock \emph{arXiv preprint arXiv:2308.12284}.

\bibitem[{Toneva et~al.(2018)Toneva, Sordoni, Combes, Trischler, Bengio, and Gordon}]{toneva2018empirical}
Mariya Toneva, Alessandro Sordoni, Remi Tachet~des Combes, Adam Trischler, Yoshua Bengio, and Geoffrey~J Gordon. 2018.
\newblock An empirical study of example forgetting during deep neural network learning.
\newblock \emph{arXiv preprint arXiv:1812.05159}.

\bibitem[{Torres et~al.(2021)Torres, Pereira, and Amini}]{torres2021survey}
Luis~Caicedo Torres, Luiz~Manella Pereira, and M~Hadi Amini. 2021.
\newblock A survey on optimal transport for machine learning: Theory and applications.
\newblock \emph{arXiv preprint arXiv:2106.01963}.

\bibitem[{Touvron et~al.(2023)Touvron, Lavril, Izacard, Martinet, Lachaux, Lacroix, Rozi{\`e}re, Goyal, Hambro, Azhar et~al.}]{touvron2023llama}
Hugo Touvron, Thibaut Lavril, Gautier Izacard, Xavier Martinet, Marie-Anne Lachaux, Timoth{\'e}e Lacroix, Baptiste Rozi{\`e}re, Naman Goyal, Eric Hambro, Faisal Azhar, et~al. 2023.
\newblock Llama: Open and efficient foundation language models.
\newblock \emph{arXiv preprint arXiv:2302.13971}.

\bibitem[{Wang et~al.(2021{\natexlab{a}})Wang, Pan, Liu, Chen, Qiu, Zhou, Huang, Chen, Lin, and Cai}]{wang2021mell}
Chengyu Wang, Haojie Pan, Yuan Liu, Kehan Chen, Minghui Qiu, Wei Zhou, Jun Huang, Haiqing Chen, Wei Lin, and Deng Cai. 2021{\natexlab{a}}.
\newblock Mell: Large-scale extensible user intent classification for dialogue systems with meta lifelong learning.
\newblock In \emph{Proceedings of the 27th ACM SIGKDD conference on knowledge discovery \& data mining}, pages 3649--3659.

\bibitem[{Wang et~al.(2021{\natexlab{b}})Wang, Thompson, and Iyyer}]{wang2021phrase}
Shufan Wang, Laure Thompson, and Mohit Iyyer. 2021{\natexlab{b}}.
\newblock Phrase-bert: Improved phrase embeddings from bert with an application to corpus exploration.
\newblock \emph{arXiv preprint arXiv:2109.06304}.

\bibitem[{Wang et~al.(2023)Wang, Chen, Ge, Xia, Bao, Zheng, Zhang, Gui, and Huang}]{wang2023orthogonal}
Xiao Wang, Tianze Chen, Qiming Ge, Han Xia, Rong Bao, Rui Zheng, Qi~Zhang, Tao Gui, and Xuanjing Huang. 2023.
\newblock Orthogonal subspace learning for language model continual learning.
\newblock \emph{arXiv preprint arXiv:2310.14152}.

\bibitem[{Wang et~al.(2021{\natexlab{c}})Wang, Chen, and Zhu}]{wang2021survey}
Xin Wang, Yudong Chen, and Wenwu Zhu. 2021{\natexlab{c}}.
\newblock A survey on curriculum learning.
\newblock \emph{IEEE Transactions on Pattern Analysis and Machine Intelligence}, 44(9):4555--4576.

\bibitem[{Wang et~al.(2022{\natexlab{a}})Wang, Kordi, Mishra, Liu, Smith, Khashabi, and Hajishirzi}]{wang2022self}
Yizhong Wang, Yeganeh Kordi, Swaroop Mishra, Alisa Liu, Noah~A Smith, Daniel Khashabi, and Hannaneh Hajishirzi. 2022{\natexlab{a}}.
\newblock Self-instruct: Aligning language model with self generated instructions.
\newblock \emph{arXiv preprint arXiv:2212.10560}.

\bibitem[{Wang et~al.(2022{\natexlab{b}})Wang, Mishra, Alipoormolabashi, Kordi, Mirzaei, Arunkumar, Ashok, Dhanasekaran, Naik, Stap et~al.}]{wang2022super}
Yizhong Wang, Swaroop Mishra, Pegah Alipoormolabashi, Yeganeh Kordi, Amirreza Mirzaei, Anjana Arunkumar, Arjun Ashok, Arut~Selvan Dhanasekaran, Atharva Naik, David Stap, et~al. 2022{\natexlab{b}}.
\newblock Super-naturalinstructions: Generalization via declarative instructions on 1600+ nlp tasks.
\newblock \emph{arXiv preprint arXiv:2204.07705}.

\bibitem[{Wang et~al.(2020)Wang, Mehta, P{\'o}czos, and Carbonell}]{wang2020efficient}
Zirui Wang, Sanket~Vaibhav Mehta, Barnab{\'a}s P{\'o}czos, and Jaime Carbonell. 2020.
\newblock Efficient meta lifelong-learning with limited memory.
\newblock \emph{arXiv preprint arXiv:2010.02500}.

\bibitem[{Wei et~al.(2021)Wei, Bosma, Zhao, Guu, Yu, Lester, Du, Dai, and Le}]{wei2021finetuned}
Jason Wei, Maarten Bosma, Vincent~Y Zhao, Kelvin Guu, Adams~Wei Yu, Brian Lester, Nan Du, Andrew~M Dai, and Quoc~V Le. 2021.
\newblock Finetuned language models are zero-shot learners.
\newblock \emph{arXiv preprint arXiv:2109.01652}.

\bibitem[{Xu et~al.(2023{\natexlab{a}})Xu, Sun, Zheng, Geng, Zhao, Feng, Tao, and Jiang}]{xu2023wizardlm}
Can Xu, Qingfeng Sun, Kai Zheng, Xiubo Geng, Pu~Zhao, Jiazhan Feng, Chongyang Tao, and Daxin Jiang. 2023{\natexlab{a}}.
\newblock Wizardlm: Empowering large language models to follow complex instructions.
\newblock \emph{arXiv preprint arXiv:2304.12244}.

\bibitem[{Xu et~al.(2023{\natexlab{b}})Xu, Tang, Shi, Zhang, Yang, Chen, and Wei}]{xu2023continual}
Zihao Xu, Xuan Tang, Yufei Shi, Jianfeng Zhang, Jian Yang, Mingsong Chen, and Xian Wei. 2023{\natexlab{b}}.
\newblock Continual learning via manifold expansion replay.
\newblock \emph{arXiv preprint arXiv:2310.08038}.

\bibitem[{Yin et~al.(2023)Yin, Liu, Yin, Zhong, Bansal, Han, and Chang}]{yin2023dynosaur}
Da~Yin, Xiao Liu, Fan Yin, Ming Zhong, Hritik Bansal, Jiawei Han, and Kai-Wei Chang. 2023.
\newblock Dynosaur: A dynamic growth paradigm for instruction-tuning data curation.
\newblock \emph{arXiv preprint arXiv:2305.14327}.

\bibitem[{Zhang et~al.(2020)Zhang, Zhang, Ghosh, Li, Tasci, Heck, Zhang, and Kuo}]{zhang2020class}
Junting Zhang, Jie Zhang, Shalini Ghosh, Dawei Li, Serafettin Tasci, Larry Heck, Heming Zhang, and C-C~Jay Kuo. 2020.
\newblock Class-incremental learning via deep model consolidation.
\newblock In \emph{Proceedings of the IEEE/CVF Winter Conference on Applications of Computer Vision}, pages 1131--1140.

\bibitem[{Zhang et~al.(2022)Zhang, Wang, and Yang}]{zhang2022continual}
Yanzhe Zhang, Xuezhi Wang, and Diyi Yang. 2022.
\newblock Continual sequence generation with adaptive compositional modules.
\newblock \emph{arXiv preprint arXiv:2203.10652}.

\bibitem[{Zhou et~al.(2023)Zhou, Liu, Xu, Iyer, Sun, Mao, Ma, Efrat, Yu, Yu et~al.}]{zhou2023lima}
Chunting Zhou, Pengfei Liu, Puxin Xu, Srini Iyer, Jiao Sun, Yuning Mao, Xuezhe Ma, Avia Efrat, Ping Yu, Lili Yu, et~al. 2023.
\newblock Lima: Less is more for alignment.
\newblock \emph{arXiv preprint arXiv:2305.11206}.

\end{thebibliography}

\appendix

\section{Appendix}
\label{sec:appendix}

\subsection{InsTag Process}
\label{sec:InsTag}

Follow \citet{lu2023instag}, we use the prompt shown in Table~\ref{tab:tagging_prompt} to employ GPT-4, providing fine-grained intention tags for given queries.
To make the word format and granularity consistent, we filter the noise in raw tags as the following steps:
\begin{itemize}
\setlength{\topsep}{0pt}
\setlength{\itemsep}{0pt}
\setlength{\parsep}{0pt}
\setlength{\parskip}{0pt}
\item \textbf{Rule Aggregation}: We replace all special characters with spaces and transform words into lowercase. Then, we apply lemmatization via NLTK~\cite{bird2009natural} to unify tag formats.
\item \textbf{Semantic Aggregation}: We obtain semantic embeddings of tags through PHRASEBERT~\cite{wang2021phrase}, a BERT-based model designed for embedding phrases. Then, we cluster tags with semantic similarity via the DBSCAN algorithm~\cite{hahsler2019dbscan}. Here, we calculate the cosine similarity and set the cluster threshold to 0.1.
\end{itemize}

\begin{table}[h]
    \centering
    \begin{tabular}{p{7cm}}
    \toprule
    You are a tagging system that provides useful tags for instruction intentions to distinguish instructions for a helpful AI assistant. Below is an instruction:\\

    [begin]
    
    \{instruction\}
    
    [end]
    
    Please provide coarse-grained tags, such as "Spelling and Grammar Check" and "Cosplay", to identify main intentions of the above instruction.
    Your answer should be a list including titles of tags and a brief explanation of each tag.
    Your response has to strictly follow this JSON format: [\{"tag": str, "explanation": str\}].
    Please respond in English. \\
    \bottomrule
    \end{tabular}
    \caption{Prompt template for annotating intention tags of the given instruction.}
    \label{tab:tagging_prompt}
\end{table}

\subsection{Data Composition}
\label{sec:Data_Composition}
SuperNI~\cite{wang2022super} collects diverse NLP tasks with instructions using the Apache-2.0 license.
The dataset curates task data in independent files, starting with a unique task ID (e.g., task001\_quoref\_question\_generation.json).
We integrate 765 English tasks from SuperNI into 16 categories, representing corresponding task IDs for each category in Table~\ref{tab:data_split}. Noting that, following the same evaluation protocol as in~\citet{wang2022super,shi2023specialist,shi2024thorough}, we adopt greedy search with a maximum generation length of 512.

\subsection{Forgetting Category Annotation}
\label{sec:FG_category_detail}
We invite 5 Chinese graduate students whose research field is related to NLP as annotation volunteers, manually labeling forgetting instances with Instruction-Related or Instruction-Unrelated.
Additionally, we have procured approval from the annotator for utilizing the data in scientific research.
We randomly sampled 3000 forgetting instances from 15 previous tasks for annotation (200 instances per task).
To better understand the forgetting category, we demonstrate detailed cases and relevant explanations in Table~\ref{tab:FG_category}.






 
 


\begin{center}
\begin{table*}[t]
\renewcommand\arraystretch{1.2}
\centering
\begin{tabular}{p{4cm}p{1cm}p{9.5cm}}
\bottomrule
\textbf{Category}& \textbf{Size} & \textbf{Task ID}\\
\hline 
Classification & 633k & 20, 50, 65, 66, 69, 70, 109, 112, 114, 115, 116, 141, 142, 143, 145, 146, 147, 148, 149, 150, 155, 190, 199, 200, 201, 202, 226, 232, 233, 242, 274, 276, 280, 290, 291, 298, 340, 341, 342, 343, 345, 346, 347, 349, 350, 351, 364, 375, 379, 382, 391, 392, 393, 400, 428, 429, 430, 431, 457, 458, 459, 472, 495, 496, 514, 515, 516, 520, 521, 564, 566, 573, 575, 577, 583, 584, 590, 614, 617, 623, 625, 629, 630, 632, 633, 638, 640, 641, 642, 679, 681, 682, 738, 767, 827, 828, 848, 854, 855, 856, 890, 907, 908, 925, 935, 936, 937, 970, 1167, 1168, 1196, 1197, 1198, 1199, 1200, 1201, 1202, 1203, 1204, 1205, 1206, 1207, 1208, 1209, 1210, 1211, 1212, 1213, 1214, 1215, 1216, 1285, 1288, 1308, 1336, 1344, 1347, 1354, 1385, 1386, 1387, 1388, 1393, 1418, 1429, 1434, 1439, 1442, 1488, 1489, 1495, 1505, 1516, 1529, 1541, 1548, 1549, 1554, 1559, 1560, 1573, 1583, 1584, 1592, 1593, 1599, 1612, 1615, 1624, 1640, 1645, 1705, 1712 \\
\hline
Generation & 506k & 1, 23, 25, 26, 59, 60, 67, 68, 71, 72, 74, 81, 82, 102, 103, 105, 166, 167, 182, 184, 191, 193, 219, 220, 246, 269, 270, 277, 278, 283, 287, 288, 294, 299, 300, 301, 303, 311, 381, 389, 405, 418, 453, 454, 455, 461, 470, 471, 489, 492, 500, 510, 547, 560, 563, 565, 568, 569, 574, 576, 581, 585, 592, 594, 599, 602, 610, 611, 619, 631, 639, 649, 672, 677, 739, 743, 760, 769, 821, 845, 847, 853, 857, 859, 860, 861, 871, 886, 897, 901, 917, 919, 927, 928, 957, 963, 964, 965, 967, 1152, 1153, 1154, 1155, 1156, 1157, 1158, 1159, 1161, 1217, 1325, 1326, 1339, 1342, 1356, 1358, 1359, 1360, 1379, 1381, 1383, 1398, 1400, 1407, 1409, 1508, 1509, 1519, 1540, 1566, 1567, 1580, 1582, 1585, 1586, 1590, 1594, 1598, 1600, 1602, 1603, 1609, 1631, 1657, 1659, 1660, 1665, 1703, 1704, 1711, 1713, 1714, 1728, 1729, 1730 \\
\hline
Program Execution & 433k & 62, 63, 64, 78, 79, 91, 93, 94, 95, 96, 97, 98, 99, 100, 101, 113, 122, 123, 124, 125, 157, 158, 159, 160, 161, 162, 163, 205, 206, 207, 208, 243, 244, 245, 267, 365, 366, 367, 368, 369, 370, 371, 372, 373, 374, 376, 377, 378, 488, 497, 499, 504, 505, 506, 507, 509, 523, 600, 605, 606, 622, 636, 637, 755, 756, 850, 851, 852, 1087, 1088, 1089, 1148, 1150, 1151, 1188, 1189, 1190, 1194, 1315, 1316, 1331, 1404, 1405, 1406, 1443, 1444, 1445, 1446, 1542, 1551 \\
\hline
Open QA & 302k & 2, 24, 28, 61, 75, 80, 83, 84, 144, 151, 152, 153, 154, 170, 194, 247, 302, 309, 310, 339, 344, 380, 390, 460, 469, 490, 491, 580, 582, 591, 595, 596, 597, 598, 615, 740, 741, 742, 745, 750, 751, 752, 753, 754, 820, 835, 849, 858, 861, 862, 863, 864, 865, 866, 867, 868, 870, 887, 898, 918, 1135, 1286, 1293, 1296, 1327, 1382, 1399, 1412, 1419, 1420, 1421, 1422, 1423, 1424, 1431, 1520, 1564, 1565, 1581, 1601, 1608, 1656, 1661, 1678, 1726, 1727, 1731 \\

\bottomrule
\end{tabular}
\end{table*}
\end{center}

\begin{center}
\begin{table*}[t]
\renewcommand\arraystretch{1.2}
\centering
\begin{tabular}{p{4cm}p{1cm}p{9.5cm}}
\bottomrule
\textbf{Category}& \textbf{Size} & \textbf{Task ID}\\
\hline
Sentiment Analysis &173k& 195, 196, 284, 285, 293, 363, 397, 398, 399, 475, 476, 477, 478, 493, 494, 512, 517, 518, 586, 587, 588, 746, 761, 819, 823, 833, 843, 844, 875, 888, 889, 902, 903, 923, 929, 1292, 1310, 1311, 1312, 1313, 1338, 1361 \\
\hline 
Comprehension & 149k & 27, 33, 44, 46, 133, 168, 176, 192, 223, 227, 248, 249, 295, 304, 329, 330, 384, 401, 403, 462, 579, 593, 648, 673, 834, 846, 891, 892, 893, 899, 900, 966, 1289, 1294, 1328, 1366, 1369, 1390, 1391, 1664 \\
\hline
Detection & 147k & 22, 88, 89, 108, 137, 209, 279, 286, 316, 317, 318, 319, 320, 321, 322, 323, 324, 325, 326, 327, 328, 333, 335, 337, 353, 354, 355, 356, 357, 358, 359, 386, 387, 513, 607, 608, 609, 904, 905, 1346, 1502, 1503, 1504, 1604, 1605, 1606, 1607, 1706, 1720, 1721, 1722, 1723, 1724, 1725 \\
\hline
Rewriting & 87k & 34, 35, 45, 111, 121, 132, 177, 275, 402, 413, 442, 550, 626, 627, 628, 670, 671, 770, 933, 934, 955, 1195, 1340, 1345, 1364, 1368, 1401, 1557, 1562, 1622, 1669, 1670 \\
\hline
Code & 71k & 76, 77, 107, 110, 126, 127, 128, 129, 130, 131, 210, 211, 212, 868, 869, 956 \\
\hline
Closed QA & 66k & 47, 73, 104, 118, 119, 138, 139, 140, 156, 164, 165, 178, 228, 229, 268, 296, 297, 385, 664, 665, 666, 667, 685, 686, 687, 688, 689, 690, 691, 692, 693, 694, 695, 696, 697, 698, 699, 700, 701, 702, 703, 704, 705, 706, 707, 708, 709, 710, 711, 712, 713, 714, 715, 716, 717, 718, 719, 720, 721, 722, 723, 724, 725, 726, 727, 728, 729, 730, 731, 732, 733, 734, 735, 736, 737, 906, 909, 1378, 1380, 1389 \\
\hline
Misc. & 66k & 43, 169, 183, 305, 306, 307, 308, 383, 567, 921, 922, 924, 1146, 1147, 1149, 1191, 1192, 1193, 1314, 1317, 1318, 1319, 1320, 1321, 1322, 1332, 1333, 1403, 1425, 1426, 1427, 1428, 1498, 1507, 1595, 1596 \\
\hline
Extraction & 59k & 36, 39, 179, 180, 181, 281, 292, 388, 456, 578, 613, 620, 645, 683, 684, 874, 926, 1447, 1448, 1449, 1451, 1452, 1453, 1479, 1480, 1481, 1482, 1483, 1484, 1485, 1486, 1487, 1506, 1510, 1517, 1518, 1568 \\
\hline
Summarization & 40k & 522, 589, 618, 668, 672, 1290, 1291, 1309, 1355, 1499, 1553, 1572 \\
\hline
Dialogue & 30k & 362, 766, 879, 880, 1384, 1394, 1500, 1501, 1531, 1533, 1534 \\
\hline
Mathematics & 24k & 85, 87, 90, 92 \\
\hline
Text Quality Evaluation & 20k & 616, 674, 675, 1186, 1283, 1284, 1341 \\

\bottomrule
\end{tabular}
\caption{\label{tab:data_split}
We analyze the intention of instructions, reclassifying the task types into 16 categories. The task IDs contained in each category are reported.
}
\end{table*}
\end{center}

\begin{center}
\begin{table*}[t]
\renewcommand\arraystretch{1.2}
\centering
\begin{tabular}{p{10cm}|p{4cm}}
\bottomrule
\textbf{Case}& \textbf{Explanation}\\
\hline 

 In this task, you are given a context tweet, a question and corresponding answer of given question. Your task is to classify given passage into two categories: (1) "Yes" if the given context is useful in answering the question, and (2) "No" if the given context is not useful. Context: \dots
\newline Ground Truth: \colorbox[RGB]{231,230,230}{No}
\newline Instruction-Related Output: \colorbox[RGB]{188,205,233}{Yes}
\newline Instruction-Unrelated Output: \colorbox[RGB]{249,198,198}{None}
& 
For close-domain instruction, we consider output within the specified range as instruction-related and vice versa as instruction-unrelated.
\\
\hline
 Craft one correct answer to the question given in input. In your answer, use as few words as possible from the given context. Use a response that is uncommon/non-stereotypical so that it is less predictable. Context: \dots, Question: \dots
\newline Ground Truth: \colorbox[RGB]{231,230,230}{He is my boyfriend.}
\newline Instruction Related Output: \colorbox[RGB]{188,205,233}{We have a close relationship.}
\newline Instruction Unrelated Output: \colorbox[RGB]{249,198,198}{10}
&
For open-domain instruction, we consider output that is relevant to the input as instruction-related, and vice versa as instruction-unrelated.
\\
\hline
 Given a command in a limited form of natural language, provide the correct sequence of actions that executes the command to thus navigate an agent in its environment. A command can be broken down into many different actions. \dots There are only six actions: 'I\_LOOK', 'I\_WALK', 'I\_RUN', 'I\_JUMP', 'I\_TURN\_LEFT', and 'I\_TURN\_RIGHT'. 
\newline jump opposite left and run opposite left.
\newline Ground Truth: \colorbox[RGB]{231,230,230}{I\_TURN\_LEFT I\_TURN\_LEFT I\_JUMP I\_TU-}
\newline \colorbox[RGB]{231,230,230}{RN\_LEFT I\_TURN\_LEFT I\_RUN}
\newline Instruction Related Output: \colorbox[RGB]{188,205,233}{I\_JUMP I\_TURN\_LEFT}
\newline Instruction Unrelated Output: \colorbox[RGB]{249,198,198}{turn left twice}
&
For the instruction that imposes restrictions on the format (e.g., within 20 words / return in the form of / should be separated with a new line / \dots), we consider output with the specified format as instruction-related, and vice versa as instruction-unrelated.
\\
\hline
Given a factoid/trivia type question, generate the topic of the question. The topic is the entity the question talks about.
\newline For which paper was reporter Clark Kent/Superman employed?
 \newline Ground Truth: \colorbox[RGB]{231,230,230}{superman, clark kent}
\newline Instruction Related Output: \colorbox[RGB]{188,205,233}{paper}
\newline Instruction Unrelated Output: \colorbox[RGB]{249,198,198}{planet}
&
For Comprehension and Summarization tasks, we consider output containing the phrases extracted from the context as instruction-related, and vice versa as instruction-unrelated.
\\
\hline
In this task, you will be given a list of integers. You should find the maximum absolute difference between 2 integers in the list. The absolute difference is the absolute value of one integer subtracted by another. The output should be a single integer which is the largest possible absolute distance.
\newline[-73, -93, -11, 79, -11, -17, -16, -52, -42, -28]
 \newline Ground Truth: \colorbox[RGB]{231,230,230}{172}
\newline Instruction Related Output: \colorbox[RGB]{188,205,233}{170}
\newline Instruction Unrelated Output: \colorbox[RGB]{249,198,198}{[-11, -17, -16]} or \colorbox[RGB]{249,198,198}{999999}
&
For tasks involving mathematical operations, we consider reasonable output in the same format as instruction-related, and vice versa as instruction-unrelated.
\\
\bottomrule
\end{tabular}
\caption{\label{tab:FG_category}
We demonstrate representative cases of two categories for a better understanding.
}
\end{table*}
\end{center}

\end{document}